\documentclass{article}

 \usepackage[final, nonatbib]{nips_2016}

\usepackage[utf8]{inputenc} 
\usepackage[T1]{fontenc}    
\usepackage{hyperref}       
\usepackage{url}            
\usepackage{booktabs}       
\usepackage{amsfonts}       
\usepackage{nicefrac}       
\usepackage{microtype}      
\usepackage{bm}		
\usepackage{amsmath}	

\title{Towards multiple kernel principal component analysis for integrative analysis of tumor samples}

\author{
  Nora K.~Speicher \\
  Max Planck Institute for Informatics\\
  Saarland Informatics Campus\\
  D-66123 Saarbr\"{u}cken, Germany\\
  \texttt{nora@mpi-inf.mpg.de} \\
   \And
  Nico Pfeifer \\
  Max Planck Institute for Informatics\\
  Saarland Informatics Campus\\
  D-66123 Saarbr\"{u}cken, Germany\\
  \texttt{npfeifer@mpi-inf.mpg.de} \\
}

\begin{document}

\maketitle

\begin{abstract}

Personalized treatment of patients based on tissue-specific cancer subtypes has strongly increased the efficacy of the chosen therapies. Even though the amount of data measured for cancer patients has increased over the last years, most cancer subtypes are still diagnosed based on individual data sources (e.g. gene expression data). We propose an unsupervised data integration method based on kernel principal component analysis. Principal component analysis is one of the most widely used techniques in data analysis. Unfortunately, the straight-forward multiple-kernel extension of this method leads to the use of only one of the input matrices, which does not fit the goal of gaining information from all data sources. Therefore, we present a scoring function to determine the impact of each input matrix. The approach enables visualizing the integrated data and subsequent clustering for cancer subtype identification. Due to the nature of the method, no free parameters have to be set. We apply the methodology to five different cancer data sets and demonstrate its advantages in terms of results and usability.

\end{abstract}

\section{Introduction}

In recent years, the amount of data that is available for cancer patients
has increased largely, both in the number of features as
well as in the number of different platforms used. One challenge that goes
along with this mass of multidimensional data (i.e. data describing different molecular levels of the tumor), is how to integrate and visualize it in a
comprehensive and sensitive manner.

In the context of data integration, multiple kernel learning provides a useful framework to optimize a weight
for each input data type. In many applications, this optimization led to better results than approaches that give equal weights to the different data
sources.
In order to visualize biological data, multiple kernel learning has been used in combination with
different dimensionality reduction schemes \cite{Lin:2011}.

However, kernel principal component analysis (kPCA), which is one of the
most widely used dimensionality reduction algorithms,  is not easily
extended using multiple kernel learning. The approach is based on the
directions of maximum variance in the data and benefits from several advantages: kPCA does not suffer from the out-of-sample problem, one does not need to fix parameters that determine a neighborhood as in local dimensionality reduction techniques like \textit{locality preserving projections}, but due to the use of the kernel function, the method provides enough flexibility to model different types of data. Although kernel PCA can be implemented in the graph embedding framework~\cite{Yan:2007}, due to an ill-posed eigenvalue problem multiple kernel PCA cannot be solved using the extended framework presented in \cite{Lin:2011}.
To be able to use this algorithm with multidimensional data, we introduce a scoring
function for the optimization of the kernel weights before one applies kPCA to the ensemble kernel matrix. The results of subsequent $k$-means clustering of the projected data show that the presented method offers some advantages compared
to naive kPCA approaches.

In the remainder of this abstract, we will shortly introduce multiple kernel learning and kPCA, followed by our
approach to combine these two concepts. We applied the method to five cancer data sets which
we will introduce before discussing the results.

\section{Methods}

\subsection{Multiple kernel learning}
In general, multiple kernel learning describes the optimization of weights $\{\beta_1, ..., \beta_M\}$ for a fixed set of input kernel matrices $\{K_1, ...K_M\}$ according
to their importance~\cite{Gonen:2011}. The aim is to find an optimal ensemble kernel matrix $\bm{K}$, which is a weighted linear combination of
the individual input kernel matrices,
\begin{equation}
    \label{eq:mkl}
    \bm{K}~=~\sum_{m=1}^M \beta_m K_m, ~~~~ \beta_m \geq 0 ~~~\text{and}~~ \sum_{m=1}^M \beta_m = 1.
\end{equation}
In this specific setting, each kernel matrix can be used to represent one data type. Due to the variety of available
kernel functions, data with different characteristics can be included, for instance quantitative data from gene expression measurements, or sequences from genome sequencing approaches.

\subsection{Kernel principal component analysis}
PCA is a global dimensionality reduction approach, which uses the
directions of maximum variance in the centered data. Having a data matrix $X$ with data points $x_i$, the first principal component is found by optimizing
\begin{equation}
    \label{eq:pca}
  \underset{\bm{\alpha}}{\mbox{arg max }}~ \sum_{i=1}^N \| \bm{\alpha}^T x_i \|^2, ~~~~ \| \bm{\alpha} \| =1.
\end{equation}
The solution is the eigenvector corresponding to the largest eigenvalue of the sample covariance
matrix, subsequent principal components are calculated analogously.
In the kernelized version, the data is implicitly projected into some
(higher dimensional) feature space using a mapping function~$\phi: x_i \rightarrow \phi(x_i)$ with $k(x_i, x_j) = \langle\phi (x_i), \phi (x_j)\rangle$ where $k$ is the positive semi-definite kernel function~\cite{Scholkopf:1999}. The directions of
maximum variance are identified in the feature space, which is achieved by considering the largest eigenvalues
of the kernel matrix and their respective eigenvectors.

\subsection{Extending kernel principal component analysis}
An extension of kPCA using multiple kernel learning would enable the user to visualize data from several sources in combination.
However, the direct implementation, i.e., 
\begin{multline}
    \label{eq:mkpca}
~~~~~~~~~~~~~~~~~~~~~~~~~~~~~~~~~~~~~~~~~~~~~~~\underset{\bm{\alpha, \beta}}{\mbox{arg max }}~ \sum_{i=1}^N \| \bm{\alpha}^T \sum_{m=1}^M \beta_m K_m \|^2,\\
\| \bm{\alpha} \| = 1; ~~~
  \beta_m \geq 0,~m = 1, ..., M ;~~~ \sum_{m=1}^M \beta_m = 1~~~~~~~~~~~~~~~~~~~~~~~~~~~~~~~~
\end{multline}
does not allow for data integration. This becomes clear when looking at Thompson's inequality concerning the eigenvalues of sums of matrices~\cite{Zhang:MatTheory}.
Consider $A$ and $B$ being $n \times n$ Hermitian matrices and $C=A+B$, with their respective eigenvalues $\lambda(A)_i$, $\lambda(B)_i$ and $\lambda(C)_i$ sorted decreasingly. Then, for any $p \geq1$,
\begin{equation}
    \label{eq:thompson}
    \sum_{i=1}^{p} \lambda(C)_i ~\leq~ \sum_{i=1}^{p} \lambda(A)_i ~+~ \sum_{i=1}^{p} \lambda(B)_i
\end{equation}
holds.
Including the kernel weight $\beta_1$ with $C~=~\beta_1 A + (1-\beta_1) B$~and~$0 \leq \beta_1 \leq 1$, we obtain the following inequality
\begin{equation}
    \label{eq:thompson2}
    \sum_{i=1}^{p} \lambda(C)_i ~\leq~ \beta_1 \sum_{i=1}^{p} \lambda(A)_i ~+~ (1-\beta_1) \sum_{i=1}^{p} \lambda(B)_i.
\end{equation}
One can see, the the right hand side is maximized if the kernel matrix with the highest sum of the $p$ largest eigenvalues has a weight of $1$. In that setting, the right hand side is equal to the left hand side. The extension to more than two kernel matrices can be made recursively. Therefore, optimizing Problem~\ref{eq:mkpca} leads to weight vectors $\bm{\beta}$ with $\beta_i = 1$ and $\beta_j = 0$ for all $j\neq i$, where $i$ is the index of the matrix with the $p$ largest eigenvalues.

Although this behavior maximizes the variance, it might not be the best choice for biological data, where we assume, that different data types can give complementing information and should therefore be considered jointly. Hence, in the following, we will introduce a scoring function, that combines the idea of kPCA with the assumption of different data supplementing each other.

\subsection{Scoring function}
Besides preserving the global variance, the main goal of this approach is to be able to integrate data from different sources that can complement each other. So, for integrating two different kernel matrices $A$ and $B$ to an ensemble kernel matrix $\bm{K}$, we propose the following gain function:
\begin{equation}
    \label{eq:gainFct}
    g_i = \text{exp}\left(\frac{\lambda(\bm{K})_i}{\text{max}(\lambda(A)_i, \lambda(B)_i, 1)} -1 \right)
\end{equation}

for each dimension $i$. Then the overall score for a projection into a $p$-dimensional space is calculated as $G = \frac{1}{p} \sum_{i=1}^{p} g_i$. The main idea is that we define a baseline, i.e., $\text{max}\{\lambda(A)_i, \lambda(B)_i, 1\}$, that represents the variance we can have by using only one matrix. Gains of variance in comparison to this baseline have a strong positive impact on the score while losses of variance are penalized only slightly. Thereby, we can account for the fact that small losses of variance in one direction often do not change the global structure of the data, but allow for more variance in a different direction. Additionally, we ensure that the baseline is not smaller than 1, which is the variance each direction would have in case of equal distribution. This scoring function is maximized to find the best kernel weights $\beta$ and can easily be extended to integrate more than two matrices.

\subsection{Materials}
We applied the approach to five different cancer sets from TCGA~\cite{TCGA}, which were preprocessed by Wang et al.~\cite{Wang:2014}. The data sets are breast invasive carcinoma (BIC; 105 samples), colon adenocarcinoma (COAD; 92 samples), glioblastoma multiforme (GBM; 213 samples), kidney renal clear cell carcinoma (KRCCC; 122 samples), and lung squamous cell carcinoma (LSCC; 106 samples). For all cancer types, gene expression, DNA methylation, and miRNA expression measurements, as well as survival data are available. The first three data types are used in the dimensionality reduction and clustering process, the latter is used to perform survival analysis in the evaluation step. Each data type is represented by a kernel matrix generated using the Gaussian kernel function. The kernel width parameter $\gamma$ was chosen according to the rule of thumb $\gamma = \nicefrac{1}{2d^2}$, with $d$ being the number of features in the matrix~\cite{Gaertner:2002}.

\section{Results}

For each cancer type, we generated results using a two-step procedure:
First, we ran the dimensionality reduction approach in order to integrate the three data types and reduce the noise in the final projection. In other approaches, the number of projection dimensions is usually determined either using the \textit{elbow method} or based on a chosen threshold for the remaining variance. Here, we benefit from the scoring function, which indicates if we gain variance in comparison to using only one matrix. Since this function does not have a global maximum, we use its first local maximum to determine the number of projection directions. Thereby, we avoid adding directions with no gain in combined variance.

In order to be able to evaluate the dimensionality reduction, we clustered the projected samples using $k$-means in the second step. The number of clusters was determined using the silhouette width~\cite{Rousseeuw:1987} of all results from 2 to 15 clusters. For each cancer type, we evaluated the resulting clusterings by comparing the survival of the patients among the different groups using the log rank test of the Cox regression model~\cite{Hosmer:2011}.
For comparison, we also used the average kernel for kPCA and the kernel with the highest variance in the first $p$ dimensions. The chosen number of dimensions and the $p$-values of the survival analysis for all three approaches can be seen in Table~\ref{table:survival}.

\begin{table}[!htbp]
\centering
\caption{Survival analysis of clustering results of kPCA used with an integrated kernel (gain function PCA), the kernel with the largest variance in the first $p$ dimensions (max variance kPCA) and an average kernel (average kPCA). In brackets, the number of clusters determined by the silhouette value are given.\label{table:survival}}{

  \begin{tabular}{lp{2.0cm}llllll}
    \toprule
    Cancer Type& number of dimensions $p$ &\multicolumn{2}{c}{gain function kPCA} & \multicolumn{2}{c}{max variance kPCA} & \multicolumn{2}{c}{average kPCA}\\
    \midrule
    BIC&3& 7.08E-3&(4)	& 0.59 &(2) 	& 5.69E-4 &(4) \\
    COAD&2& 6.47E-3&(2)	& 6.47E-3 &(2)	&  3.28E-2 &(3)\\
    GBM&2& 0.79&(3)	& 5.29E-2 &(13)	& 1.52E-2 &(8)\\
    KRCCC&3& 8.53E-3&(15)	& 2.54E-2 &(14)	& 1.15E-2 &(8) \\
    LSCC&3& 7.52E-3&(3)	& 7.52E-3& (3)	& 9.22E-3 &(3)\\
    \bottomrule
    \end{tabular}}{}    
\end{table}

The number of dimensions $p$ determined by the scoring function was for all cancer types rather small, either two or three, which can be due to the fact that we used three input data types. However, as we can see in the survival analysis, this projection allowed the identification of biologically significant clusters within the cancer types.

Using the conservative significance threshold of $p \leq 0.01$, we can see that our method was able to find significant clusters in all cancer types but GBM, while both other methods identified significant clusters only for two out of the five cancer types.
In the GBM data, the gene expression kernel is very dominant in terms of variance, therefore, it obtains a high weight. However, there is no clear group structure in this matrix, as can also be seen from the high number of clusters that were obtained by max variance kPCA, and neither this method nor gain function PCA is able to find a clustering that correlates with the survival of the patients. For KRCCC, there is a very small group of patients with different survival behavior which we find at the expense of having 15 clusters. In general, the results for the LSCC data are very stable; for all other cancer types, at least one of the naive approaches results in a clustering with no significant difference in survival times between the patient groups.

\section{Conclusion and outlook}
In this work, we presented a data integration method based on kernel principal component analysis. We showed that the direct extension of kPCA for several data sources does not allow for data integration. Thus, we proposed a scoring function to determine the best combination of the input data. On five cancer data sets, we showed that this procedure works in most cases better or as good as naive approaches in terms of survival analysis. Additionally, our method alleviates the user from the choice of free parameters, since the number of dimensions to project into can be determined by the introduced scoring function. New samples from the same cancer type can be easily projected into the learnt subspace to observe similarities in the neighborhood.

Besides the survival data, the clusters could also vary in other clinical aspects (e.g., response to treatment). Investigating on this in combination with an analysis of the molecular foundation of the clusterings, for instance the identification of differentially methylated sites or differentially expressed genes, could reveal beneficial insights concerning the molecular mechanisms in tumor cells and consequently their treatment.

\bibliographystyle{unsrt}
\bibliography{nips_2016}{}

\end{document}